\documentclass[journal]{IEEEtran}
\usepackage[utf8]{inputenc}
\ifCLASSINFOpdf
\usepackage[pdftex]{graphicx}
\usepackage{graphics}
\else
\fi

\usepackage[cmex10]{amsmath}
\usepackage{amssymb}  
\usepackage[numbers,sort&compress]{natbib}

\usepackage{algorithmic}

\usepackage{array}
\usepackage{multirow}

\usepackage{url}

\usepackage{hyperref}

\begin{document}
%
\title{Progressive LiDAR Adaptation for Road Detection}
%
%
%

\author{
Zhe~Chen,
Jing~Zhang,
and~Dacheng~Tao,~\IEEEmembership{Fellow,~IEEE}
\thanks{
\textit{Corresponding author: Jing Zhang}.

This work was supported by Australian Research Council Projects FL-170100117, DP-180103424, IH-180100002, and National Natural Science Foundation of China (NSFC) under Grant 61806062.

Z. Chen, and D. Tao are with the UBTECH Sydney Artificial Intelligence Centre and the School of Computer Science, in the Faculty of Engineering and Information Technologies, at the University of Sydney, 6 Cleveland St, Darlington, NSW 2008, Australia (email: zche4307@uni.sydney.edu.au; dacheng.tao@sydney.edu.au).

J. Zhang is a visiting scholar with the School of Software and Advanced Analytics Institute, at the University of Technology Sydney, 15 Broadway, Ultimo NSW 2007, Australia. He is a lecturer with School of Automation at the Hangzhou Dianzi University (email: jing.zhang@uts.edu.au).

\textcopyright 2018 IEEE. Personal use of this material is permitted.
Permission from IEEE must be obtained for all other uses, in any current or future media, including reprinting/republishing this material for advertising or promotional purposes, creating new collective works, for resale or redistribution to servers or lists, or reuse of any copyrighted component of this work in other works. 
}
}

%
%

\markboth{IEEE/CAA JOURNAL OF AUTOMATICA SINICA,~Vol.~X, No.~X, X~X}%
{Progressive LiDAR Adaptation for Road Detection}

%



\maketitle
\begin{abstract}
Despite rapid developments in visual image-based road detection, robustly identifying road areas in visual images remains challenging due to issues like illumination changes and blurry images. To this end, LiDAR sensor data can be incorporated to improve the visual image-based road detection, because LiDAR data is less susceptible to visual noises. However, the main difficulty in introducing LiDAR information into visual image-based road detection is that LiDAR data and its extracted features do not share the same space with the visual data and visual features. Such gaps in spaces may limit the benefits of LiDAR information for road detection. To overcome this issue, we introduce a novel Progressive LiDAR Adaptation-aided Road Detection (PLARD) approach to adapt LiDAR information into visual image-based road detection and improve detection performance. In PLARD, progressive LiDAR adaptation consists of two subsequent modules: 1) data space adaptation, which transforms the LiDAR data to the visual data space to align with the perspective view by applying altitude difference-based transformation; and 2) feature space adaptation, which adapts LiDAR features to visual features through a cascaded fusion structure. 
Comprehensive empirical studies on the well-known KITTI road detection benchmark demonstrate that PLARD takes advantage of both the visual and LiDAR information, achieving much more robust road detection even in challenging urban scenes. In particular, PLARD outperforms other state-of-the-art road detection models and is currently top of the publicly accessible benchmark leader-board.
\end{abstract}


\IEEEpeerreviewmaketitle

\section{Introduction}

Robust urban road detection is critical for autonomous driving systems. Without adequate recognition of road areas, a self-driving vehicle could not make safe decisions to achieve reliable navigation. Over the years, segmentation techniques have been used to identify road areas in monocular images, and, more recently,
the introduction of deep convolutional neural network(DCNN)-based image segmentation methods such as FCN \cite{long2015fully} and DeepLab \cite{chen14semantic} has significantly improved the performance of visual image-based road detection.

Despite progress (e.g. \cite{shinzato2014road, kuhnl2012spatial, xiao2015crf,mendes2015vision,levi2015stixelnet,Mendes:2016, mohan2014deep,chen2017rbnet}), DCNNs may still under-perform when there are visual noises such as variable illumination, over exposure, ambiguous appearances, and blurry images. 
To overcome these issues and improve road detection performance, many studies \cite{caltagirone2018lidar, caltagirone2017fast, chen2017lidar} have introduced LiDAR information to improve road detection. ``LiDAR''\footnote{an acronym of \textit{light detection and ranging}} refers to the data acquired by measuring the distance to a target by illuminating the target with pulsed laser light\cite{lidar:online}. Many studies have proved that LiDAR is robust to various visual noises and can complement monocular image data. 
For example, Caltagirone \textit{et al.} \cite{caltagirone2017fast} reported that 3D LiDAR point clouds provide sufficient information to detect roads and are robust to visual noises, making it possible to robustly detect roads using only LiDAR data. Moreover, the authors of \cite{xiao2017hybrid} attempted to fuse LiDAR and visual information for road detection.
However, existing approaches that utilize LiDAR data for road detection are still far from effective and provide only limited improvements over visual image-based road detection methods. Here, we investigate the difficulties encountered when utilizing LiDAR data for road detection and propose a novel and more effective approach to incorporate LiDAR information into a visual image-based road detection system.

\begin{figure*}[t]
\begin{center}
\includegraphics[width=0.7\linewidth,height=0.25\textheight]{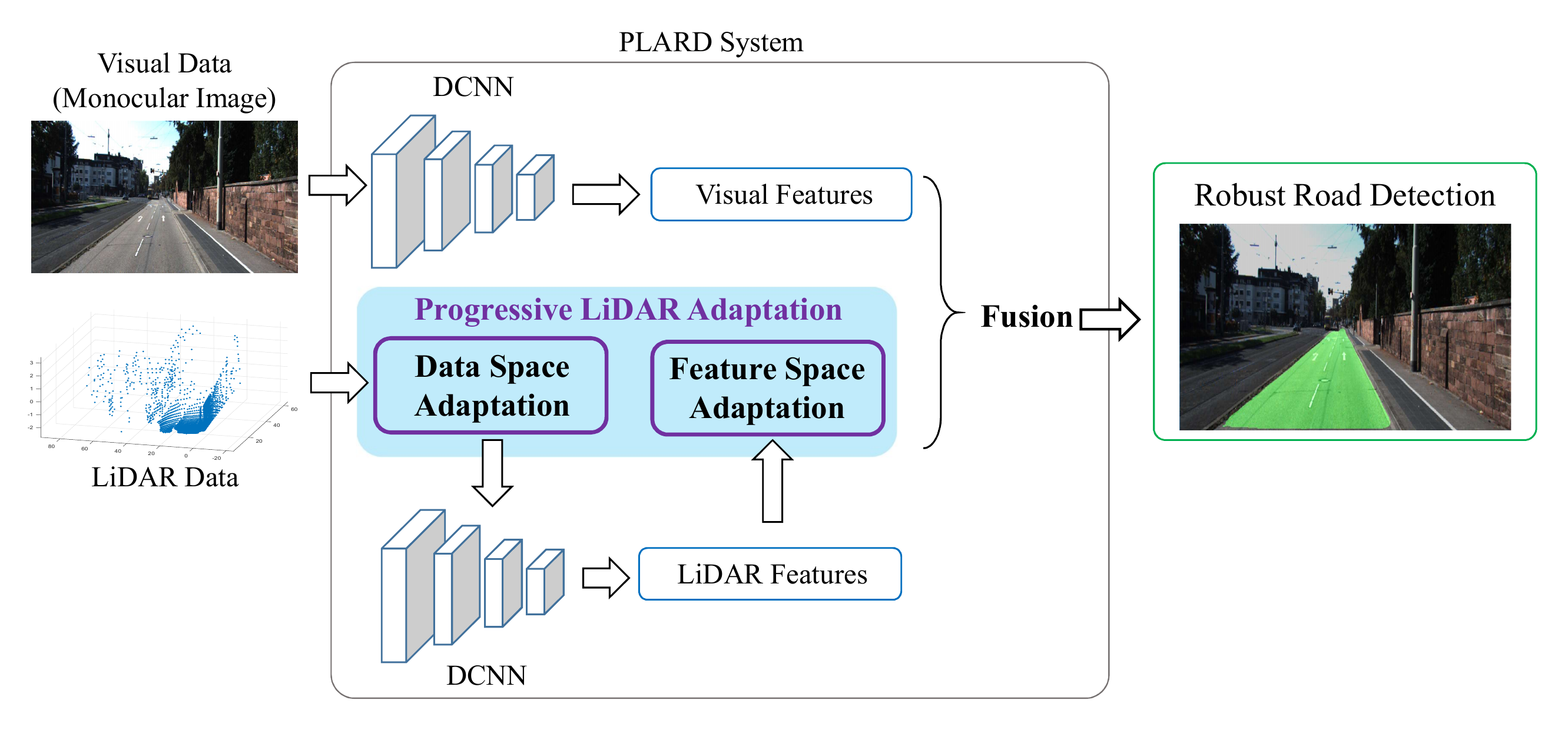}
\end{center}
   \caption{Overview of the Progressive LiDAR Adaptation-aided Road Detection (PLARD) approach. We overcome the issue that LiDAR information and visual information are in different spaces when detecting road areas in urban scenes. In particular, 
the proposed progressive LiDAR adaptation consists of a data space adaptation step that adapts the view of raw LiDAR data to align the view of visual images and a feature space adaptation step that adapts the LiDAR features to visual features. By fusing the adapted LiDAR information and the visual information, PLARD achieves robust road detection.}
\label{fig:title}
\end{figure*}

By interrogating LiDAR information and visual information for road detection, we conclude that two major factors would pose difficulties for effective cooperation between the two types of information. First, since raw LiDAR data and raw visual image data are in different spaces, it is difficult to define a proper space to integrate both data types. For instance, in the KITTI road detection dataset \cite{Geiger2012CVPR,Geiger2013IJRR}, the provided LiDAR data is defined in the 3D real-world space, while the visual images are defined on the 2D image plane. Although researchers can project the LiDAR data onto the 2D image plane using the calibration parameters, this may at the same time alter the road appearance in the LiDAR data, making road areas less distinguishable in the LiDAR data space. As a result, it will be difficult for a DCNN-based road detection model to learn a reliable road detection function based on the LiDAR data, let along improve the visual image-based road detection models. Moreover, it is also difficult to appropriately integrate the features extracted from the LiDAR data and the visual features extracted from visual images. More specifically, since the road appearance in the LiDAR data is described by scattered points and road appearance in visual data is described by RGB values of pixels on the 2D image plane, it is highly likely that the features extracted from both data sources are also in different spaces. This gap in feature spaces could adversely impact the feature fusion performance and the final detection accuracy, thus existing feature fusion methods for road detection can hardly outperform state-of-the-art visual image-based road detection algorithms.

To overcome these issues, here we propose a novel progressive LiDAR adaptation technique to make LiDAR information more compatible with visual information and to improve the visual image-based road detection more effectively. To achieve this, in the progressive LiDAR adaptation, we introduce proper transformation functions to successively adapt the LiDAR data space into the visual data space and adapt the LiDAR feature space into the visual feature space. Accordingly, the progressive LiDAR adaptation procedure consists of a data space adaptation step and a feature space adaptation step. The data space adaptation step transforms and aligns the LiDAR data space to the visual data space whilst still making the road areas easy-to-distinguish in the LiDAR data. Afterwards, through a cascaded fusion structure, the feature space adaptation step transforms the LiDAR feature space into the space that better complements and improves the visual features. By integrating the visual information with the adapted LiDAR information, we obtain a more robust road detection model: Progressive LiDAR Adaptation-aided Road Detection (PLARD) model. Fig. \ref{fig:title} shows an overview of our proposed system.

Using the well-known KITTI road detection benchmark \cite{Geiger2012CVPR}, we perform comprehensive experiments for the proposed PLARD system to evaluate the effectiveness of different parts of the proposed technique and the overall performance gains over a visual image-based road detection system. Empirical results demonstrate that LiDAR information delivers more benefits for road detection via our proposed progressive LiDAR adaptation technique. Furthermore, on the test set of KITTI road detection benchmark, PLARD promisingly improves the road detection accuracy, outperforming other visual image-based road detection algorithms, LiDAR-based road detection algorithms, and the algorithms that fuse both information. In particular, PLARD achieves state-of-the-art performance on the publicly accessible leader-board.
Indeed, an ensemble of 3 our PLARD models ranks the top on the leader-board at the time of writing this paper.

\begin{figure*}[t]
\begin{center}
\includegraphics[width=\linewidth,height=0.27\textheight]{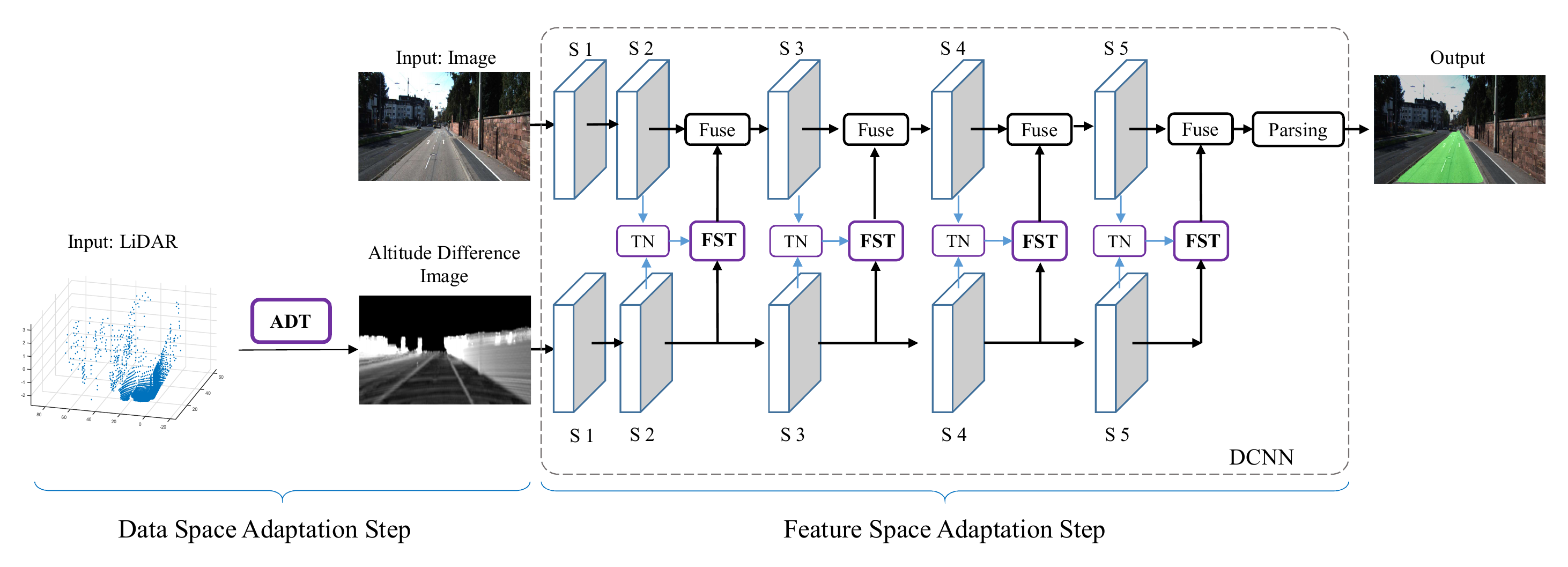}
\end{center}
   \caption{Overall pipeline of the proposed PLARD system. In the data space adaptation step, we introduce the Altitude Difference-based Transformation (denoted ``\textbf{ADT}'') to adapt the raw LiDAR data, obtaining a better aligned LiDAR data space where the roads are easier to be distinguished from other objects. In the feature space adaptation step, DCNNs are first employed to detect roads on visual and LiDAR data respectively. Then, Feature Space Transformation (denoted ``\textbf{FST}'') modules are introduced to transform the LiDAR features and make them better complement and improve visual features. In each ``\textbf{FST}'' module, a Transformation Network (denoted ``{TN}'') is employed to learn the transformation parameters. After feature transformation, the visual features and the adapted LiDAR features are fused via a cascaded structure. The cascaded fusion integrates features at all the convolution stages (denoted ``{S1-S5}'') but the first stage. Lastly, in the parsing stage, PLARD performs classification on the integrated features, delivering robust road detection results.}
\label{fig:main}
\end{figure*}

\section{Related Work}
Road detection is beneficial to various other autonomous tasks \cite{qi2018dynamic,8405356,kong2009vanishing, chen2017dynamically,chen2017generic,chen2018context}. 
Over the years, various algorithms have been developed to tackle the road detection problem \cite{xing2018advances, han2018semi,munoz2017deep,munoz2010stacked}. For instance, model-based methods build shape \cite{aly2008real,laddha2016map} or appearance models \cite{alvarez2013learning} to describe the road structure and then identify road areas in the input images. Learning-based methods then attempt to employ classifiers (such as SVM \cite{zhou2010road} and random forest \cite{xiao2016monocular}) to distinguish road from non-road areas. In practice, learning-based methods usually perform better than model-based methods.

By considering the road detection task as a semantic segmentation task, DCNNs have been demonstrated in recent years to be particularly useful for road detection. In particular, several typical algorithms have proven to be effective in semantic segmentation. For example,
Long \textit{et al.}\cite{long2015fully} proposed the fully convolutional and upsampling layers to tackle the pixel-level semantic segmentation problems. Moreover, authors of \cite{chen2017deeplab,liang2015semantic} achieved compelling semantic segmentation performance by introducing dilated convolution operations which can greatly enlarge receptive fields of convolutional kernels without reducing the resolution of the feature maps. By taking advantage of both the fully convolutional layer and dilated convolution operations, several studies \cite{yu2015multi, lin2017refinenet, yu2017dilated} have achieved impressive performance on semantic segmentation benchmarks. These techniques have been widely used for detecting roads in urban scenes \cite{mohan2014deep,teichmann2016multinet,Mendes:2016,levi2015stixelnet}. 

In order to improve the effectiveness of DCNN-based road detection, several promising algorithms have been proposed. For instance, Mendes \textit{et al.} \cite{Mendes:2016} introduced a large contextual window and a network-in-network architecture to improve accuracy, while the study \cite{OB16b} introduced an efficient deep network following the ``encoder-decoder" principle as discussed in ``U-net" \cite{ronneberger2015u}.
However, DCNNs are still susceptible to visual noises and they usually require excessively long processing times to guarantee better performance. As an example, the algorithm in \cite{levi2015stixelnet} cost around 1s to process an image and the algorithm in \cite{mohan2014deep} needed 2s, and both algorithms fail to achieve state-of-the-art performance, making them impractical for moving platforms like autonomous vehicles.

Despite this progress in the visual image-based road detection, others proposed that LiDAR is robust to visual noises and they have attempted to detect the roads using LiDAR information. In \cite{kuhnl2012spatial}, visual images were transformed from the perspective view into the birds-eye-view for road detection. Another study \cite{caltagirone2017fast} took LiDAR point clouds rather than visual images as input, which can perform promising road detection in 3D real-world space. However, these studies did not effectively take advantage of both types of information, limiting their final detection performance. There are also studies \cite{xiao2017hybrid,xiao2015crf} that attempted to fuse both visual and LiDAR information for road detection, but existing fusion-based algorithms are either time-consuming or less effective than other state-of-the-art algorithms. In this study, we hypothesized that the difficulty of using LiDAR to improve road detection is due to the gaps between data spaces and feature spaces for LiDAR information and visual information. To overcome this issue, we
introduce a progressive LiDAR adaptation technique that effectively adapts and integrates LiDAR information into the visual image-based road detection pipeline to boost the robustness and accuracy for road detection.

\section{PLARD System}

\subsection{Problem Definition}
In this study, we formulate the road detection task as assigning pixels on the 2D image plane with a binary label indicating whether the pixel belongs to road areas. LiDAR information will be adapted for the road detection on the 2D image plane.
Formally, suppose $\hat{y}$ represents a ground-truth label, $f$ is a road detection function, and $W$ is the model parameters of $f$. Taking LiDAR data $L$ and visual data $I$ as input, we tackle the road detection problem by optimizing the following objective: 
\begin{equation}
\mathop{\min}_W\sum_{i}\sum_{x,y} \mathcal{L}(f(I_i, L_i; W), \hat{y}) |_{x,y},
\label{eq:def}
\end{equation}
where $i$ indexes over training examples, $x,y$ represent the horizontal and vertical offsets, respectively, on the image plane, and $\mathcal{L}$ is a loss function.

\begin{figure*}[t]
\begin{center}
\includegraphics[width=0.75\textwidth,height=0.25\textheight]{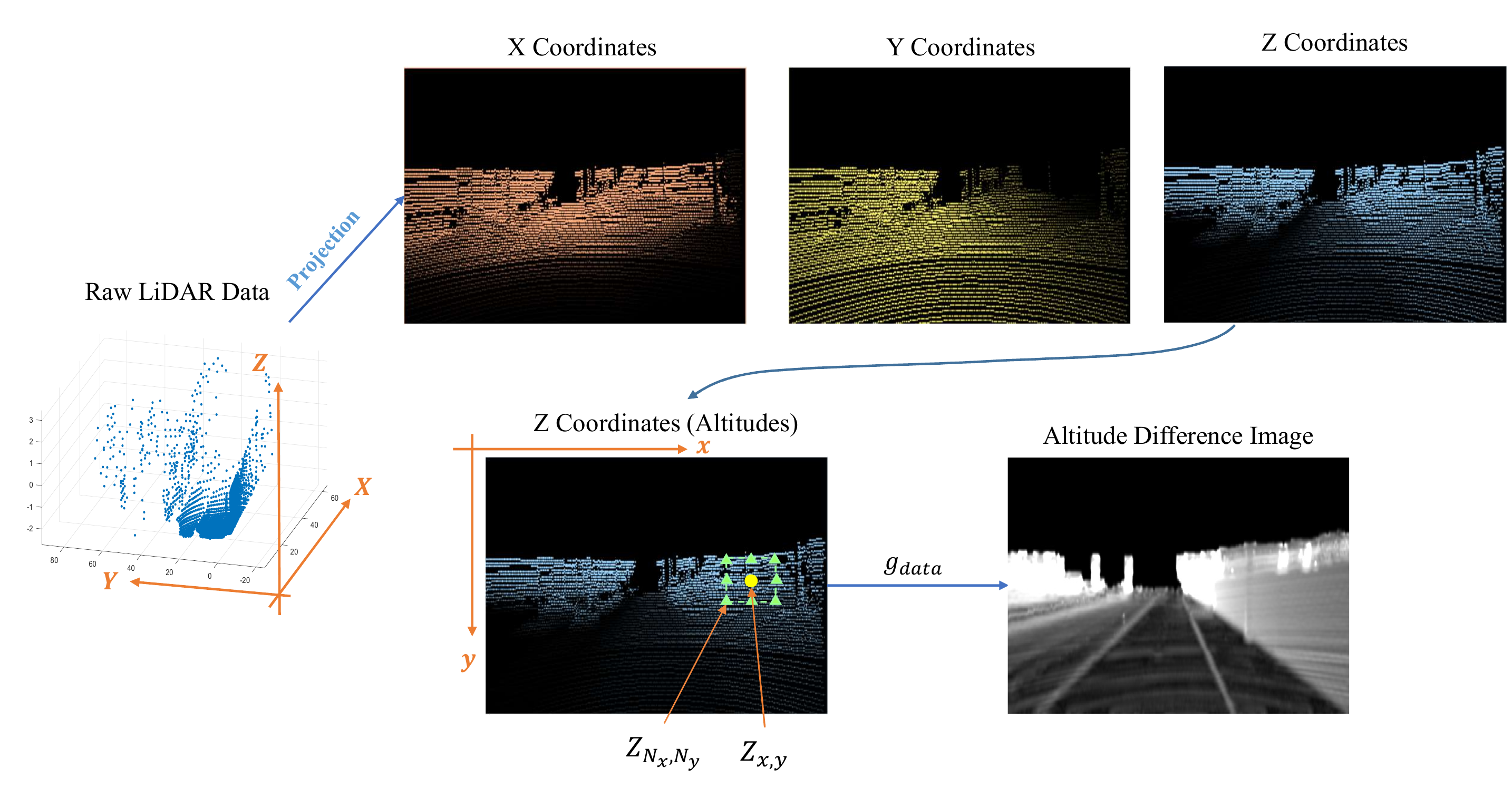}
\end{center}
   \caption{An example of directly projected LiDAR data and altitude difference image. Using provided calibration parameters, LiDAR points described by 3-dimensional coordinate vectors in the real-world space can be projected onto the 2D image plane. Upper-right figures show the projected LiDAR points whose intensities represent the normalized $X$, $Y$, and $Z$ coordinates. According to the definition of axes, $Z$ can be considered as altitudes and then we can compute the absolute values of changes in altitudes with respect to spatial offsets between two locations (such as $Z_{x,y}$ and $Z_{N_x, N_y}$ in the figure). With the computed differences in altitude, an altitude difference image (as illustrated in the bottom-right figure) is obtained. The altitude difference image can make roads easier to be distinguished from non-road areas by preserving the characteristics of road in the LiDAR data. Best view this figure in color.}
\label{fig:LiDAR}
\end{figure*}

\subsection{Overview}
Considering that gaps exist between the data and feature spaces, robust road detection is difficult to achieve solely by using a simple combination of LiDAR and visual information. To overcome this problem and improve road detection, we propose a progressive LiDAR adaptation technique to make the LiDAR information more compatible with visual information, thus more effectively incorporating both information types. 
Mathematically, we formulate that the road detection function $f$ of PLARD system takes the following form: 
\begin{equation}
f(I, L; W) = f_{parsing}(f_{fuse}(f_{vis}(I;W_{vis}), g(L; W_{lidar}))),
\label{eq:parsing}
\end{equation}
where $g$ is the progressive LiDAR adaptation function, $f_{vis}$ is the visual image-based road detection function, $W_{vis}$ and $W_{lidar}$ are parameters of the corresponding functions, $f_{fuse}$ is a fusion operation, and $f_{parsing}$ is the final binary classification function that identifies road areas from fused features. Specifically, we implement $f_{vis}$ by introducing a ResNet101\cite{he2016deep} backbone and implement $f_{parsing}$ by a 2-class softmax function after a pyramid scene parsing module \cite{zhao2017pyramid}. 

We implement the progressive adaptation function $g$ by introducing two subsequent adaptation steps: data space adaptation step and feature space adaptation step. In the data space adaptation, we transform raw LiDAR data from the 3D space to the 2D image plane while preserving the distinguishable characteristics of road areas. Then, in the feature space adaptation step, we introduce a learning-based module to transform the LiDAR features such that the transformed features better complement the visual features in road detection. 
Accordingly, we formulate the progressive LiDAR adaptation function $g$ as follows:
\begin{equation}
 g(L; W_{lidar}) = g_{feat}(f_{lidar}(g_{data}(L); W_{lidar})),
\end{equation}
where $g_{data}$ and $g_{feat}$ represent the data space adaptation function and feature space adaptation function, respectively, and $f_{lidar}$ is the LiDAR-based road detection function. The overview of PLARD is shown in Fig. \ref{fig:main}. 

In Section \ref{sec:da}, we implement $g_{data}$ by studying the altitude changes with respect to the 2D image plane. In Section \ref{sec:fa}, we implement $g_{feat}$ by introducing a learning-based module into the DCNN architecture to transform the features extracted from LiDAR data into a space that better complement the visual features. Lastly, the implementation details of the PLARD approach is described in Section \ref{sec:exp}.

\subsection{LiDAR Adaptation}
In the data space adaptation step of the progressive LiDAR adaptation, we introduce a novel altitude difference-based transformation method to transform the LiDAR data space. In the feature space adaptation step, we introduce a transformation network to learn and transform the LiDAR feature space. 

Both raw LiDAR data and raw visual data are in different spaces.
Raw LiDAR data is composed of tens of thousands of points in the 3D real-world space and each LiDAR point is described by a 3-dimensional coordinate vector, while the visual data is composed of pixels on a 2D image plane and each pixel is described by a RGB value. Therefore, it is extremely challenging to directly and smoothly integrate visual data with raw LiDAR data. 
Fortunately, with the help of calibration parameters, the 3D LiDAR points can be projected onto the 2D image plane, thereby obtaining an image with the projected LiDAR points. Although this obtained image can be used for road detection, we argue that road and non-road appearances in the projected LiDAR data will be less distinguishable from each other, thus diminishing the capacity of a road detection model to identify road areas, as illustrated in the upper-right corner of Fig. \ref{fig:LiDAR}. 
Instead of using direct projection results, we propose an altitude difference-based transformation operation that better preserves the road characteristics to instantiate $g_{data}$ and help adapt the 3D LiDAR data to the visual data space.

\subsubsection{\textbf{Data Space Adaptation}}
\label{sec:da}
In the first stage, we introduce \textit{altitude difference-based transformation} to implement data space adaptation.
Altitude difference-based transformation is introduced based on the observation that road surfaces in the 3D space are flat and relatively smooth in the altitudes of LiDAR point clouds compared to other objects such as vehicles and buildings. After projecting LiDAR points onto the image plane, such smoothness can be maintained by recording the altitudes of the original 3D LiDAR points. As a result, road areas can be better distinguished in the projected LiDAR data according to the changes in altitudes on the image plane. 

More specifically, on the LiDAR-projected 2D image plane, altitude difference-based transformation computes a pixel value $V_{x,y}$ located at $(x,y)$ according to:
\begin{equation}
g_{data}(L)|_{x,y}=V_{x,y} = \frac{1}{M} \sum_{N_x,N_y} \frac{| Z_{x,y} - Z_{N_x, N_y} |}{\sqrt{(N_x-x)^2+(N_y-y)^2}},
\label{eq:diff}
\end{equation}
where $Z(x,y)$ is the altitude of the LiDAR point projected on $(x,y)$, $(N_x, N_y)$ denote positions in the neighbourhood of $(x,y)$, and $M$ is the total number of considered neighborhood positions. If a neighboring pixel is not correlated to a 3D point, we simply ignore it. It is worth mentioning that Eq. \ref{eq:diff} can be viewed as a mean absolute value for gradients of altitudes of projected points with respect to the 2D image plane. Therefore, if an object is upright and sharp, its projected areas will have large altitude differences on the image plane. For example, in Fig. \ref{fig:LiDAR}, road areas commonly have a small intensity on the altitude difference image while other objects generally have large intensities, distinguishing the roads from other objects. Suppose $H$ and $W$ are the height and width of the input image, respectively. According to Eq. \ref{eq:diff}, the complexity of altitude difference-based transformation is $O(MHW)$.

\subsubsection{\textbf{Feature Space Adaptation}}
\label{sec:fa}
In addition to the gap in data spaces, the features extracted from LiDAR data could also be inconsistent with the visual features extracted from images, since road areas may have different appearances in different data sources. This feature space inconsistency could further hamper the performance of integrating LiDAR features and visual features, thus limiting the overall benefit of introducing LiDAR information. We therefore attempt to transform the LiDAR feature space to make the LiDAR features better complement and improve visual features and visual image-based road detection performance. 
However, the main challenge of feature space adaptation is that we do not have a complete knowledge about how to properly transform the feature space. To tackle this issue, we introduce a learning-based module to find an appropriate space adaptation operation. 

In general, we assume that the linear transformation can properly define a feature space adaptation operation, and we have:
\begin{equation}
g_{feat}(\mathbf{f}_{lidar}) = \mathbf{\alpha} \mathbf{f}_{lidar} + \mathbf{\beta},
\end{equation} 
where $\mathbf{\alpha}$ is a scalar vector, $\mathbf{\beta}$ is an offset vector, and $\mathbf{f}_{lidar}$ is the LiDAR feature to be adapted: 
\begin{equation}
\mathbf{f}_{lidar} = f_{lidar}(g_{data}(L); W_{lidar}).
\label{eq:lidar}
\end{equation}

To estimate $\mathbf{\alpha}$ and $\mathbf{\beta}$ properly and achieve better feature space adaptation, we introduce a neural network, called the transformation network, to learn and adapt LiDAR features.  
Accordingly, the transformation network estimates $\mathbf{\alpha}$ and $\mathbf{\beta}$ following:
\begin{eqnarray}
\mathbf{\alpha} = f_{\alpha}(\mathbf{f}_{lidar}, \mathbf{f}_{vis}; W_{\alpha}), \\
\mathbf{\beta} = f_{\beta}(\mathbf{f}_{lidar}, \mathbf{f}_{vis}; W_{\beta}),
\end{eqnarray}
where $f_{\alpha}$ and $f_{\beta}$ represent the neural network function for computing $\mathbf{\alpha}$ and $\mathbf{\beta}$, respectively, $W_{\alpha}$ and $W_{\beta}$ are their corresponding weight parameters, and $\mathbf{f}_{vis}$ is the visual feature:
\begin{equation}
\mathbf{f}_{vis} = f_{vis}(I; W_{vis}).
\end{equation}

\begin{figure}[t]
\begin{center}
\includegraphics[width=\linewidth,height=0.25\textheight]{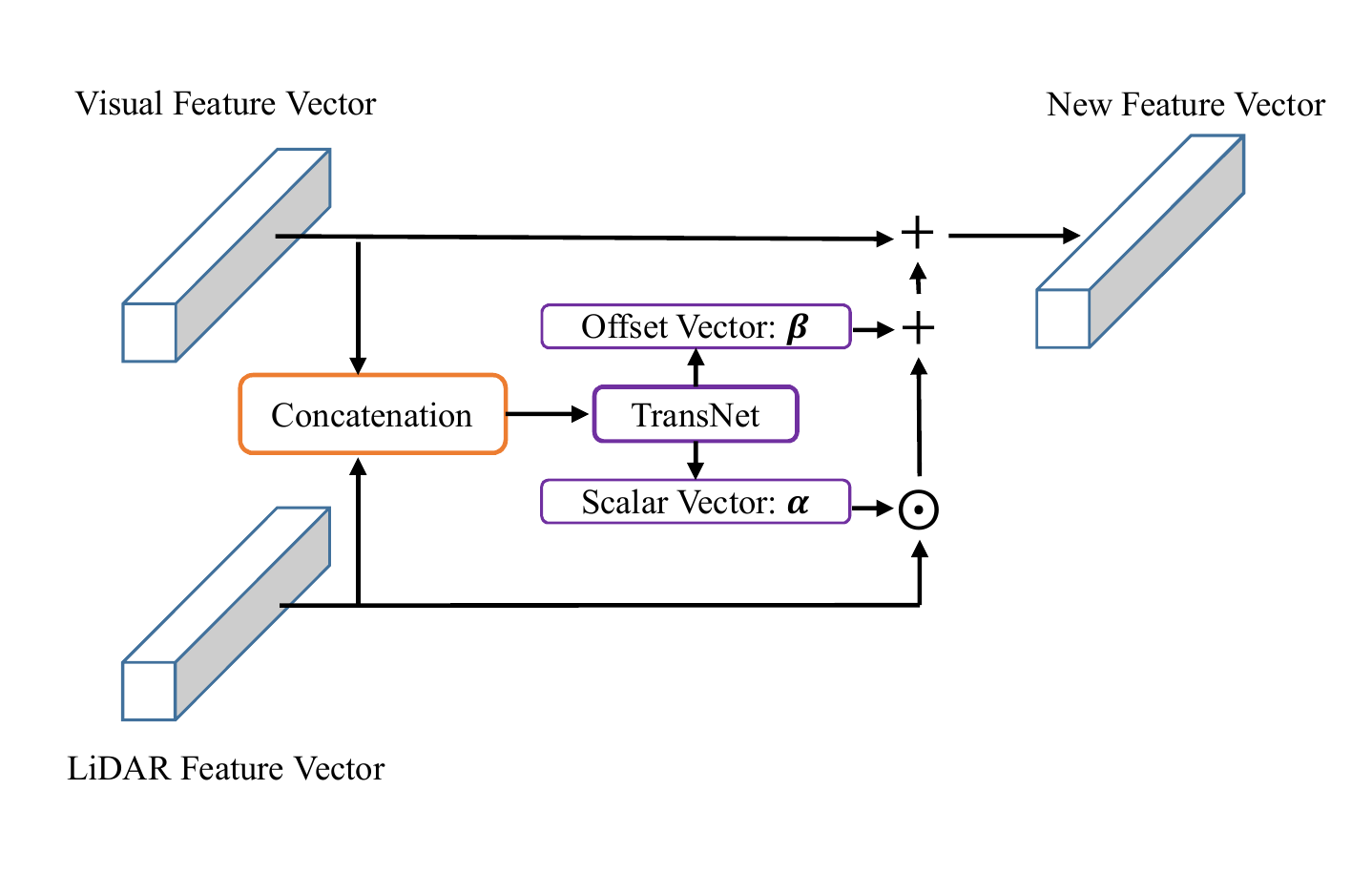}
\end{center}
   \caption{Structure of a feature adaptation module in PLARD. Accepting as input the convolutional features of visual information and LiDAR information, the feature space adaptation introduces a transformation network (denoted ``TransNet'') to learn and transform LiDAR features. More specifically, the transformation network outputs a scalar vector $\mathbf{\alpha}$ and an offset vector $\mathbf{\beta}$ based on the concatenated visual and LiDAR features. By fusing the adapted LiDAR features with visual features via a residual structure, we obtain an improved feature vector. In this figure, $\odot$ means element-wise multiplication and $+$ means addition in this figure.}
\label{fig:fa}
\end{figure}

In a DCNN, we implement both $f_{\alpha}$ and $f_{\beta}$ by using fully convolutional operations whose weight parameters are defined by $W_{\alpha}$ and $W_{\beta}$, respectively. $\mathbf{f}_{lidar}$ and $\mathbf{f}_{vis}$ are concatenated as input for both $f_{\alpha}$ and $f_{\beta}$.
By optimizing the parameters $W_{\alpha}$ and $W_{\beta}$ together with the overall road detection system, we can learn a proper transformation of feature spaces for the extracted LiDAR features, thus improving road detection more promisingly. Fig. \ref{fig:fa} shows a detailed schematic of the feature space adaptation stage. Comparing to the overall DCNN architecture which may have over dozens of convolutional layers, the complexity of feature space adaptation is small because it only involves three 1x1 convolutional operations and three element-wise multiplication or addition operations. Suppose $C$ is the channel number, $H_k$ and $W_k$ are height and width of the $k$-th convolutional stage, respectively, then the complexity of feature space adaptation for that stage is the complexity of 3 convolutional operations: $O(4C^2H_k W_k)$. 

\subsection{Cascaded Fusion for Adapted LiDAR Information}
After LiDAR is adapted according to $g_{data}$ and $g_{feat}$, we fuse the LiDAR and visual information to achieve more robust road detection. Specifically , we implement the fusion function $f_{fuse}$ using an residual-based cascaded fusion structure. In this architecture, we make the adapted LiDAR features improve the visual features under a residual structure. By applying this residual-based fusion to every subsequent convolutional stage in a DCNN pipeline, we achieve robust road detection. 

Mathematically, taking visual features $\mathbf{f}_{vis}$ and the adapted LiDAR-based features $\mathbf{f}_{lidar}$ as input, we implement the fusion function as follows:
\begin{eqnarray}
f_{fuse}^k(f_{vis}^k(f_{fuse}^{k-1}(I,L);W^k_{vis}), g^k(L; W_{lidar}))\notag\\
 = f_{vis}^k(f_{fuse}^{k-1}(I,L);W^k_{vis}) +\lambda g^k(L; W_{lidar})),
 \label{eq:fuse}
\end{eqnarray}
where $k$ indicates features from the $k$-th convolutional stage of the DCNN in the road detection system and $\lambda$ is a scalar parameter. It is worth noting that there are 5 convolutional stages when using ResNet-101 \cite{he2016deep} as the backbone network. The right part of Fig. \ref{fig:main} shows the detailed structure of this fusion.

\subsection{Overall Objective}
During optimization, we train PLARD in an end-to-end manner to obtain parameters for both visual image-based and LiDAR-based DCNNs. In addition to the objective for road detection with the fused LiDAR and image information, we also introduce a loss for the LiDAR-based DCNN such that the LiDAR-based road detection system directly detects roads from altitude difference images. Moreover, we follow the design of \cite{zhao2017pyramid} and introduce an auxiliary loss in the visual image-based DCNN to facilitate convergence.

According to Eq. \ref{eq:def}, training PLARD is to minimize a loss function w.r.t. all of its parameters. Suppose $\mathcal{L}_{PLARD}$ is the overall loss for road detection with the fused LiDAR and image information, $\mathcal{L}_{LiDAR}$ represents the loss only for the LiDAR-based DCNN, and $\mathcal{L}_{aux}$ is the auxiliary loss in the visual image-based DCNN. Then, the loss function of PLARD can be written as:
\begin{equation}
\mathcal{L} = w_{parsing} \mathcal{L}_{parsing} + w_{lidar} \mathcal{L}_{lidar} + w_{aux} \mathcal{L}_{aux},
\label{eq:l}
\end{equation}
where $w_{parsing}$, $w_{lidar}$ and $w_{aux}$ are the corresponding loss weights. The settings for $w_{parsing}$, $w_{lidar}$, and $w_{aux}$ can be found in Sec. \ref{sec:impl}.

In this study, we exploit multinomial cross-entropy loss to define all these losses:
\begin{equation}
\mathcal{L}_{\{parsing,lidar,aux\}} = - \sum_{c=1}^2(\hat{y}^c~log_{10}(y^c_{\{parsing,lidar,aux\}})),
\end{equation}
where $\hat{y}^c$ is the ground-truth for category $c$ and $y^c$ is the prediction outputs. Specifically, $y^c_{parsing}$, $y^c_{lidar}$, and $y^c_{aux}$ are computed according to Eq. \ref{eq:parsing}, Eq. \ref{eq:lidar} (only for the final convolutional stage), and Eq. \ref{eq:fuse} (only for the convolutional stage $k=4$), respectively. 

\begin{table*}
\caption{Ablation studies for PLARD. Results are based on 5-fold cross validation using the training set. Four components are compared, including``img'' (visual image-based road detection), ``L-Proj'' (road detection using directly projected LiDAR data), ``L-ADT'' (road detection using altitude difference transformed LiDAR data), and ``FSA'' (road detection by applying feature space adaptation on LiDAR features through cascaded fusion). Best scores are highlighted in {\bf bold}. 
}
\label{tab:ab}
\begin{center}
\resizebox{0.9\textwidth}{!}{ 
  \begin{tabular}{c c c c| c | c c  c  c c c }
  \hline
   Img & L-Proj & L-ADT & FSA  & Speed (s/im) & Max F & AP & PRE & REC & FPR & FNR \\
  \hline
  \checkmark & &  & & 0.120 & 88.29& 91.17 & 86.98 & 89.73 & 7.62 & 10.28 \\
   & \checkmark&  & & 0.045& 83.25& 87.54 & 79.68 & 87.55 & 13.54 & 12.45 \\
   & & \checkmark& & 0.042& 86.39& 90.73 & 84.73 & 88.21 & 9.62 & 11.79 \\
   \checkmark& \checkmark&  &&0.161 & 89.34& 91.40 & 87.68 & 91.22 & 6.75 & 8.78 \\
  \checkmark & &  \checkmark& & 0.159 & 89.79& 91.78 & 89.34 & 90.36 &6.46 & 9.64 \\
  \checkmark & & \checkmark & \checkmark & 0.160& {\bf 92.85}& {\bf93.14} & {\bf93.16} &{\bf 92.55 }&{\bf 3.10} &{\bf 7.45 }\\
  \hline
  \end{tabular}
}
\label{fig:ab}
\end{center}
\end{table*}

\begin{table*}[t]
\caption{Overall performance on the test set of KITTI road detection benchmark. Best scores are highlighted in {\bf bold}. ``+'': a 3-model ensemble with multi-scale testing.}
\label{tab:all}
\begin{center}
\resizebox{\textwidth}{!}{
  \begin{tabular}{|>{\centering}p{2.5cm} | >{\centering}p{1.7cm}| >{\centering}p{2cm}| >{\centering}p{1.3cm} | c |c | c | c | c |}
  \hline
   Methods & Speed (s/im) & Input & MaxF & AP   & PRE  & REC  & FPR  & FNR \\
  \hline
  SPRAY\cite{kuhnl2012spatial} &0.045&Image & 87.09 \% & 91.12 \% & 87.10 \% & 87.08 \% & 7.10 \% & 12.92 \%\\
  FCN\_LC\cite{Mendes:2016} & 0.03&Image & 90.79 \% & 85.83 \% & 90.87 \% & 90.72 \% & 5.02 \% & 9.28 \% \\
  HybridCRF\cite{xiao2017hybrid} & 1.5 &Image + LiDAR& 90.81 \% & 86.01 \% & 91.05 \% & 90.57 \% & 4.90 \% & 9.43 \% \\
  FTP \cite{laddha2016map}&0.28 &Image& 91.61 \% & 90.96 \% & 91.04 \% & 92.20 \% & 5.00 \% & 7.80 \% \\
  Up\_Conv\cite{OB16b} &0.08&Image & 93.83 \% & 90.47 \% & 94.00 \% & 93.67 \% & 3.29 \% & 6.33 \% \\
  LoDNN\cite{caltagirone2017fast} & 0.018 &LiDAR& 94.07 \% & 92.03 \% & 92.81 \% & 95.37 \% &	4.07 \% &	4.63 \% \\
  MultiNet\cite{teichmann2016multinet} & 0.17&Image & 94.88 \% & 93.71 \% & 94.84 \% & 94.91 \% & 2.85 \% & 5.09 \% \\
  StixelNet II\cite{garnett2017real} & 1.2&Image& 94.88 \% & 87.75 \% & 92.97 \% & 96.87 \% & 4.04 \% & 3.13 \% \\
  RBNet\cite{chen2017rbnet} & 0.18&Image & 94.97 \% & 91.49 \% &	94.94 \% & 95.01 \% &	2.79 \% & 4.99 \% \\
  LidCamNet \cite{caltagirone2018lidar} & 0.15 & Image + LiDAR & 96.03 \% & 93.93 \% & 96.23 \% & 95.83 \% & 2.07 \% & 4.17 \% \\	
  NF2CNN  & 0.006 & Image + LiDAR & 96.70 \%& 89.93 \%	& 95.37 \% & {\bf98.07} \% & 2.62 \% & {\bf1.93} \% \\
  PSPNet\cite{zhao2017pyramid} & 0.12  &Image & 96.29 \% & 93.71 \% & 96.22 \% & 96.35 \% & 2.09 \% & 3.65 \% \\
  \hline
  \hline
  PLARD & 0.16 &Image + LiDAR&  96.83 \% & 93.98 \% & 96.79 \% & 96.86 \% & 1.77 \% & 3.14 \% \\
  PLARD$^+$ & 1.5 &Image + LiDAR& {\bf97.03} \% & {\bf 94.03} \% & {\bf 97.19} \% & 96.88 \% & {\bf 1.54} \% & 3.12 \% \\
  \hline
  \end{tabular}
}
\end{center}
\end{table*}

\begin{table*}
\caption{Performance on different tasks in the test set of KITTI road detection benchmark. Best scores are highlighted in {\bf bold}. ``+'': a 3-model ensemble with multi-scale testing.}
\label{tab:task}
\begin{center}
\resizebox{0.75\textwidth}{!}{ 
  \begin{tabular}{|>{\centering}p{2.5cm} | c | c | c | c | c | c |}
  \hline
   \multirow{2}{*}{Methods} & \multicolumn{2}{c|}{UM} &\multicolumn{2}{c|}{UMM} &\multicolumn{2}{c|}{UU} \\
   \cline{2-7}
   & Max F & AP & Max F & AP & Max F & AP\\
  \hline
SPRAY\cite{kuhnl2012spatial} & 88.14 \% & 91.24 \% & 89.69 \% & 93.84 \% & 82.71 \% & 87.19 \%\\
FCN\_LC\cite{Mendes:2016} & 89.36 \% & 78.80 \% & 94.09 \% & 90.26 \% & 86.27 \% & 75.37 \%\\
HybridCRF\cite{xiao2017hybrid} & 90.99 \% & 85.26 \% & 91.95 \% & 86.44 \% & 88.53 \% & 80.79 \%\\
FTP \cite{laddha2016map} & 91.20 \% & 90.60 \% & 92.98 \% & 92.89 \% & 89.62 \% & 88.93 \%\\
Up\_Conv\cite{OB16b} & 90.48 \% & 88.20 \% & 93.89 \% & 92.62 \% & 91.89 \% & 89.44 \%\\
LoDNN\cite{caltagirone2017fast} & 92.75 \% & 89.98 \% & 96.05 \% & 95.03 \% & 92.29 \% & 90.35 \%\\
MultiNet\cite{teichmann2016multinet} & 93.99 \% & 93.24 \% & 96.15 \% & 95.36 \% & 93.69 \% & 92.55 \%\\
StixelNet II\cite{garnett2017real} & 94.05 \% & 85.85 \% & 96.22 \% & 91.24 \% & 93.40 \% & 85.01 \%\\
RBNet\cite{chen2017rbnet} & 94.77 \% & 91.42 \% & 96.06 \% & 93.49 \% & 93.21 \% & 89.18 \%\\
LidCamNet \cite{caltagirone2018lidar} & 95.62 \% & {\bf 93.54} \% & 97.08 \% & 95.51 \% & 94.54 \% & 92.74 \%\\
NF2CNN & 96.09 \% & 88.40 \% & 97.77 \% & 93.31 \% & 95.47 \% & 86.98 \% \\
  PSPNet\cite{zhao2017pyramid} & 95.62 \% & 92.95 \% & 96.95 \% & 95.38 \% & 95.86 \% & 92.73 \% \\
  \hline
  \hline
  PLARD & 96.34 \% & 93.43 \% & 97.53 \% & 95.61 \% & {\bf 96.13} \% & 93.00 \% \\
  PLARD$^+$ & {\bf 97.05} \% &93.53 \% & {\bf 97.77} \% & {\bf 95.64} \% & 95.95 \% & {\bf 95.25} \% \\
  \hline
  \end{tabular}
}
\end{center}
\end{table*}
\section{Experiment}
\label{sec:exp}
In this section, we evaluate the effectiveness of PLARD on the KITTI road benchmark. 
We first adopt 5-fold cross-validation to illustrate the effectiveness of the individual parts in PLARD. Then, we evaluate PLARD on test set of the benchmark and compare it to other state-of-the-art road detection algorithms.\footnote{Results on the test set are available on: \url{http://www.cvlibs.net/datasets/kitti/eval\_road.php}.}

\subsection{Dataset}
The KITTI road benchmark \cite{Geiger2012CVPR} is popular with road detection researchers due to its comprehensiveness. KITTI uses a wide variety of evaluation metrics to assess algorithm performance and also provides information captured by various sensors including visual cameras, LiDAR sensor, and GPS. KITTI contains 289 images for training and 290 images for testing, both containing three different road scene categories including Urban Marked roads (UM), Urban Multiple Marked lanes (UMM), and Urban Unmarked roads (UU). For fair evaluation, KITTI does not provide ground-truths for test images, and the number of submissions for online evaluation is limited. All the results are publicly accessible on its official website. 

\subsection{Evaluation Metrics.}
We follow the standard evaluation metrics used by KITTI, which are detailed in  \cite{Fritsch2013ITSC}. The metrics include the maximum F1-measure (MaxF), average precision (AP), precision rate (PRE), recall rate (REC), false positive rate (FPR), and false negative rate (FNR). The four latter measures are obtained at the working point of MaxF. According to KITTI’s evaluation system, all the results are transformed into birds-eye-view space for evaluation and MaxF is selected as the metric for ranking the evaluated algorithms.

\subsection{Implementation and Model Training}
\label{sec:impl}
The implementation of PLARD is simple and straight-forward. In this section, we discuss some implementation details that should be taken into account.

When dealing with altitude difference images, we set $N_x$ and $N_y$ as positions within a $7\times7$ window centered at $(x,y)$, thus the maximum value of $M$ is 48 (excluding the centre). All the values in an altitude difference image are re-scaled within the range $[0,255]$. 

In PLARD, we employ PSPNet \cite{zhao2017pyramid} as the visual image-based DCNN and use ResNet-101 \cite{he2016deep} as the backbone. We also employ a 101-level DCNN to implement $f_{lidar}$ and extract LiDAR features.  
To avoid excessive computational loads introduced by using two DCNNs, we make the channel numbers of the LiDAR-based DCNN 8 times smaller than the channel numbers of the visual image-based DCNN at the same level. In addition, we use hybrid convolutions \cite{wang2018understanding} in the LiDAR-based DCNN to augment its expressive capacity with fewer channel numbers. During fusion, we use a uniform channel number, i.e. 256. The parameter $\lambda$ in Eq. \ref{eq:fuse} is set to 0.1. Regarding loss weights in Eq. \ref{eq:l}, we empirically set $w_{parsing}$, $w_{aux}$, and $w_{lidar}$ as 1.0, 0.16, 0.4, respectively.

We resize all the images into 384 by 1280 during training and testing. We adopt the SGD algorithm to optimize all parameters with the learning rate gradually decaying from $1\times10^{-4}$ to $1\times10^{-6}$. We train models using 80 epochs and use NVIDIA's GTX Titan GPUs for computation. 
The implementation of PLARD and the experimental settings are similar in both the ablation studies and the evaluation on the test set. The differences are as follows. First, to better illustrate the effectiveness of different parts of our model, we train models from scratch in the ablation studies. In contrast to ablation studies, we pre-train visual image-based DCNN in the PLARD system using external data \cite{MVD2017} to improve robustness for the evaluation on test set. In addition, for the test set, we improve PLARD by adopting several data augmentation techniques, including multi-scale training and testing, random cropping, and disturbing the image brightness. Lastly, we extend the training period for the test set evaluation to three times longer than in the ablation study.

\subsection{Ablation Study}
We first evaluate the effectiveness of different PLARD components and compare their performances to baseline methods using 5-fold cross-validation on the training set. Table \ref{tab:ab} shows the average results for all the ``UM'', ``UMM'', and ``UU'' tasks. In particular, we investigate (1) the benefits of altitude difference-based transformation operations compared to directly projected LiDAR points; and (2) the improvements afforded by the proposed learning-based feature space adaptation module implemented via a cascaded fusion structure compared to the direct fusion implemented via a simple concatenation.

It can be seen that both the altitude difference-based transformation technique and the feature space adaptation technique achieve promising improvements compared to their counterparts. Specifically, training with the altitude difference image (L-ADT) surpasses the road detection model using directly projected LiDAR points (L-Proj) by around 3 points for ``Max F''. In addition, fusing both visual and LiDAR information improves the visual image-based road detection baseline, indicating that LiDAR is helpful for boosting robustness. By further introducing the feature space adaptation procedure through a cascaded fusion structure, the final model (Img + L-ADT + FSA) achieves the highest performance among compared methods, demonstrating the effectiveness of the feature transformation and cascaded fusion structure. Note that the speed for LiDAR-based road detection does not include processing time for raw data projection and transformation.

\begin{figure*} 
\begin{center}
\includegraphics[width=\textwidth,height=0.35\textheight]{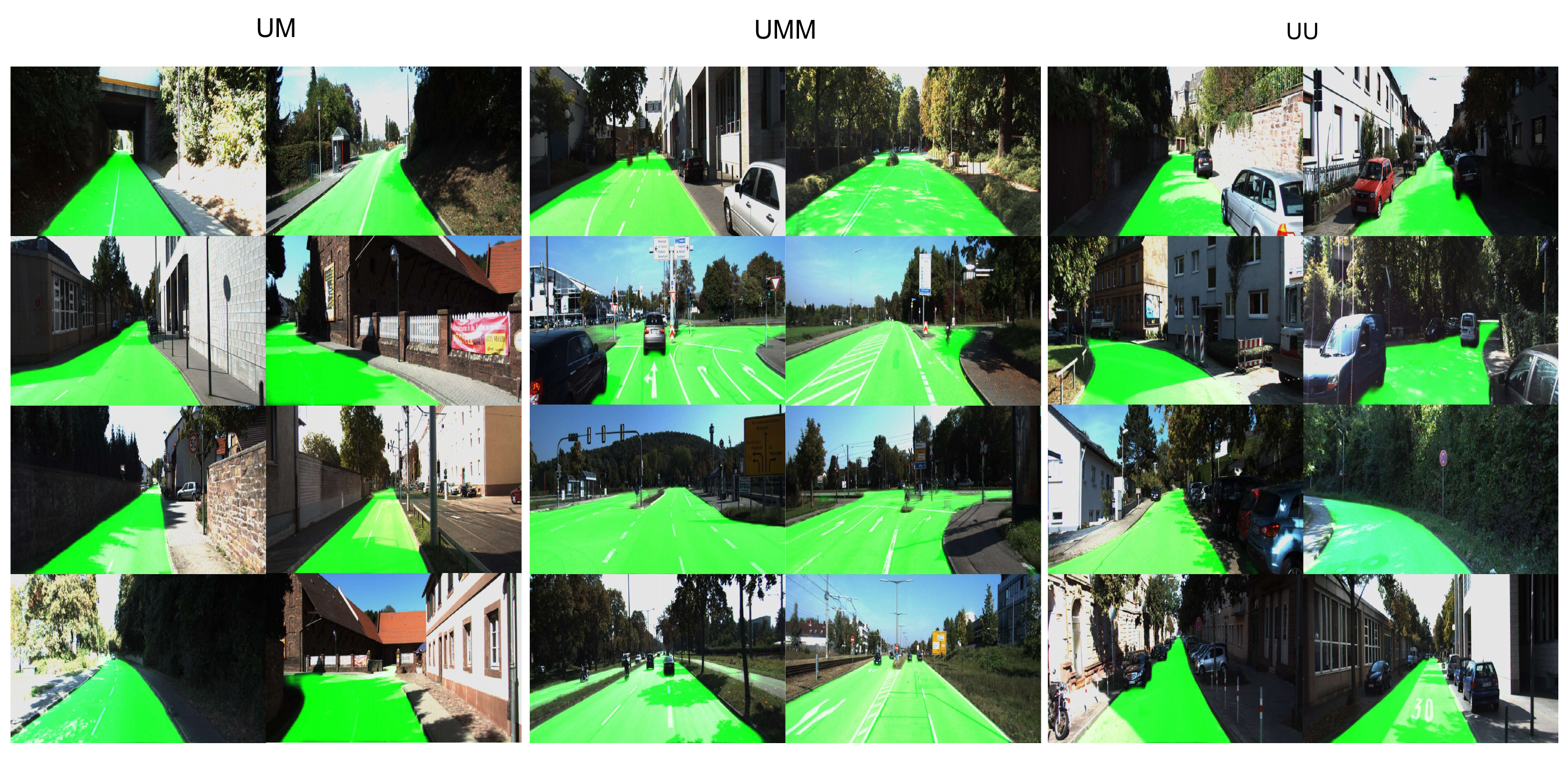}
\end{center}
   \caption{Qualitative results of PLARD in different road scenes. Presented images show that PLARD is robust to severe illumination conditions. Best view in color.}
\label{fig:qualitative}
\end{figure*}
\subsection{Evaluation on the Test Set}
\subsubsection{Quantitative Results}
Using all the training images in the benchmark, we evaluate the visual image-based PSPNet as the road detection baseline and we evaluate the performance of a single PLARD model as well as a 3-model PLARD ensemble using the multi-scale testing for augmentation. We compare the visual image-based PSPNet and our models with other state-of-the-art road detection algorithms, including SPRAY \cite{kuhnl2012spatial}, FCN-LC \cite{Mendes:2016}, HybridCRF\cite{xiao2017hybrid}, FTP \cite{laddha2016map}, Up-Conv \cite{OB16b}, LoDNN\cite{caltagirone2017fast}, MultiNet \cite{teichmann2016multinet}, Stixel-Net II\cite{garnett2017real}, RBNet\cite{chen2017rbnet}, LidCamNet \cite{caltagirone2018lidar} and NF2CNN. All results were computed on the KITTI evaluation server, and the results of other studies are based on the scores reported on KITTI's website. Overall algorithm performance is shown in Table \ref{tab:all}, and the detailed performance for different tasks, such as UM, UMM, and UU, is shown in Table \ref{tab:task}. 

The overall results of PLARD and other state-of-the-art road detection systems are illustrated in details in Table \ref{tab:all}. The single PLARD model alone promisingly improves the visual image-based PSPNet and achieves superior MaxF score compared to other road detection algorithms. By further augmenting the PLARD with multi-scale testing and model ensemble, we achieve the best scores for most of the metrics, demonstrating the effectiveness of the proposed system for robust road detection. In particular, our method achieves the highest MaxF and AP scores which are commonly used as the performance indicators of an approach.

In addition to the overall results, we also compare performance with respect to separate tasks (``UM'', ``UMM'', and ``UU'') in Table \ref{tab:task}. It can be observed that PLARD also outperforms other compared algorithms in most of the Max F and AP metrics. Especially, on ``UM'', the augmented PLARD system outperforms NF2CNN, the best-performing algorithm among other methods, by around 1 point for MaxF. Moreover, for AP on the ``UU'' task, our method surpasses the second-ranking method LidCamNet\cite{caltagirone2018lidar} by around 2.5 points, demonstrating the generalizability of the proposed PLARD system.

\subsubsection{Qualitative Results}
Fig. \ref{fig:qualitative} shows some qualitative results of PLARD on test set of the benchmark. Columns from left to right show the road detection results of PLARD on UM, UMM, and UU tasks, respectively. It can be seen that our method is robust to severe illumination conditions like heavy shadows and over-exposed areas. 

\section{Conclusions}
In this study, we introduce a novel road detection method called PLARD by progressively adapting LiDAR information to visual information. The PLARD first performs data space adaptation, which adapts LiDAR data to the 2D image space to align with the perspective view by applying an altitude difference-based transformation. Then, the PLARD performs feature space adaptation to adapt the learned LiDAR features to visual features through cascaded fusion layers. By leveraging these two adaptation modules successively, PLARD takes advantage of both the visual and LiDAR information and is robust to various challenging aspects of urban scenes. We validate the PLARD model on the well-known KITTI road detection benchmark, where it outperforms other state-of-the-art road detection models and currently ranks the top of the leader-board, demonstrating its effectiveness and superiority over existing methods.

\ifCLASSOPTIONcaptionsoff
  \newpage
\fi

\bibliographystyle{IEEEtran}
\bibliography{reference}

\begin{IEEEbiography}[{\includegraphics[width=1in,height=1.25in,clip,keepaspectratio]{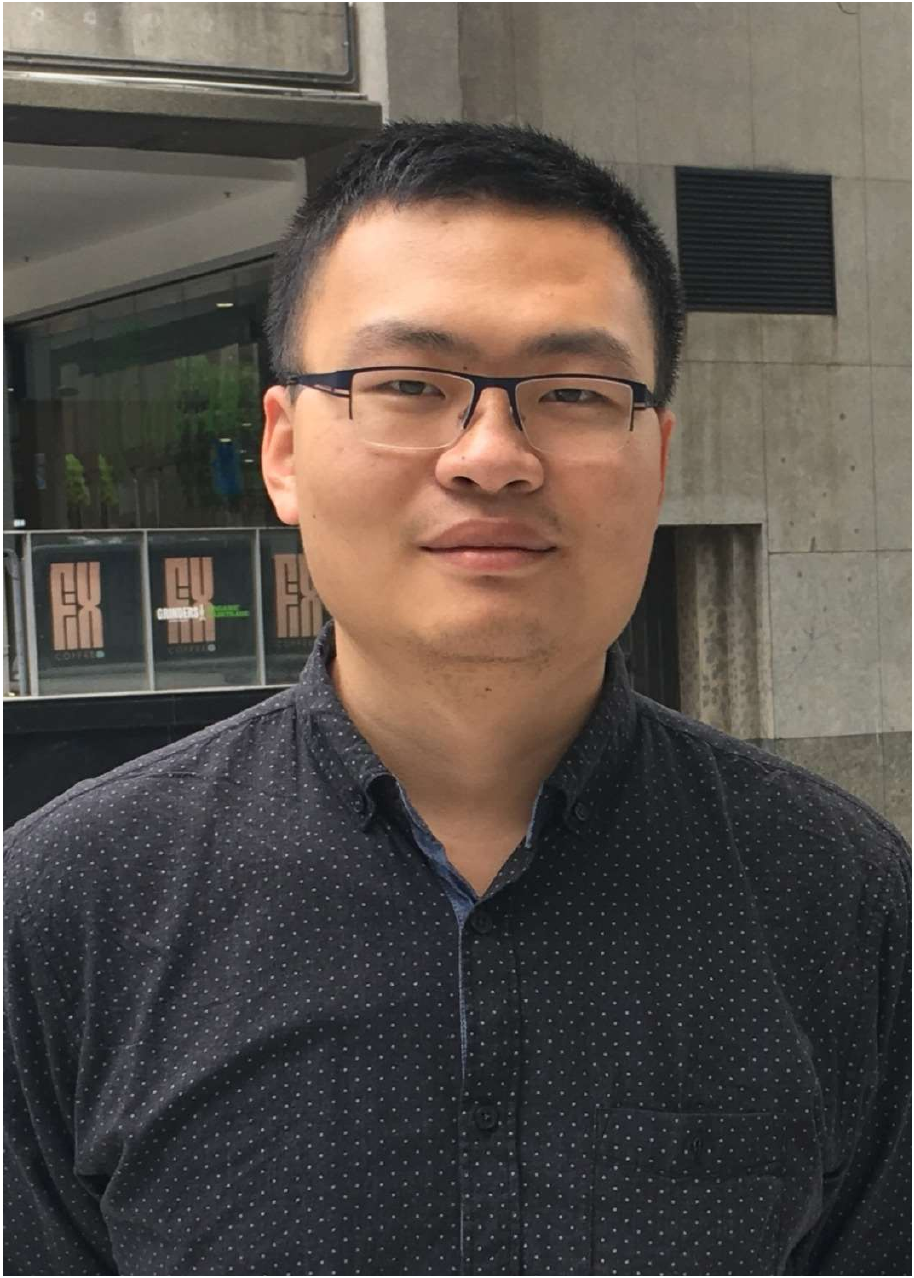}}]{Zhe Chen} received the B.S. degree in Computer Science from University of Science and Technology
of China, Hefei, China, in 2014. He is currently pursuing the Ph.D. degree at the UBTECH Sydney Artificial Intelligence Centre and the School of Computer Science, the Faculty of Engineering and Information Technologies, the University of Sydney. His current research interests
include object recognition, computer vision, and deep learning. His studies was published in IEEE CVPR, ICONIP, and ECCV. 
\end{IEEEbiography}

\begin{IEEEbiography}[{\includegraphics[width=1in,height=1.25in,clip,keepaspectratio]{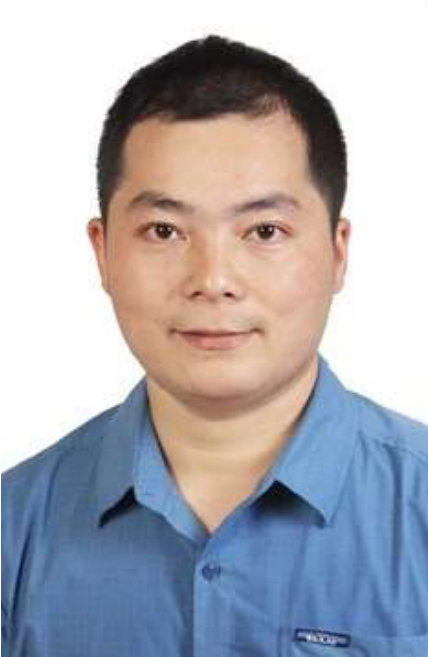}}]{Jing Zhang} received the B.S. degree from the Henan University and the Ph.D. from the University of Science and Technology of China (USTC). He used to work as a research fellow at the IFLYTEK Research. Then, he has been a lecturer at the School of Automation in the Hangzhou Dianzi University since 2017. Currently, he is a visiting scholar at the School of Software and Advanced Analytics Institute in the University of Technology Sydney. His research interests include computer vision and multimedia. He published several papers at IEEE CVPR, ACM Multimedia, AAAI, IEEE TCSVT, Information Sciences, Neurocomputing, etc. He serves as a reviewer for a number of journals and conferences such as TIP, TCSVT, Information Sciences, Neurocomputing, ACM Multimedia.
\end{IEEEbiography}

\begin{IEEEbiography}[{\includegraphics[width=1in,height=1.25in,clip,keepaspectratio]{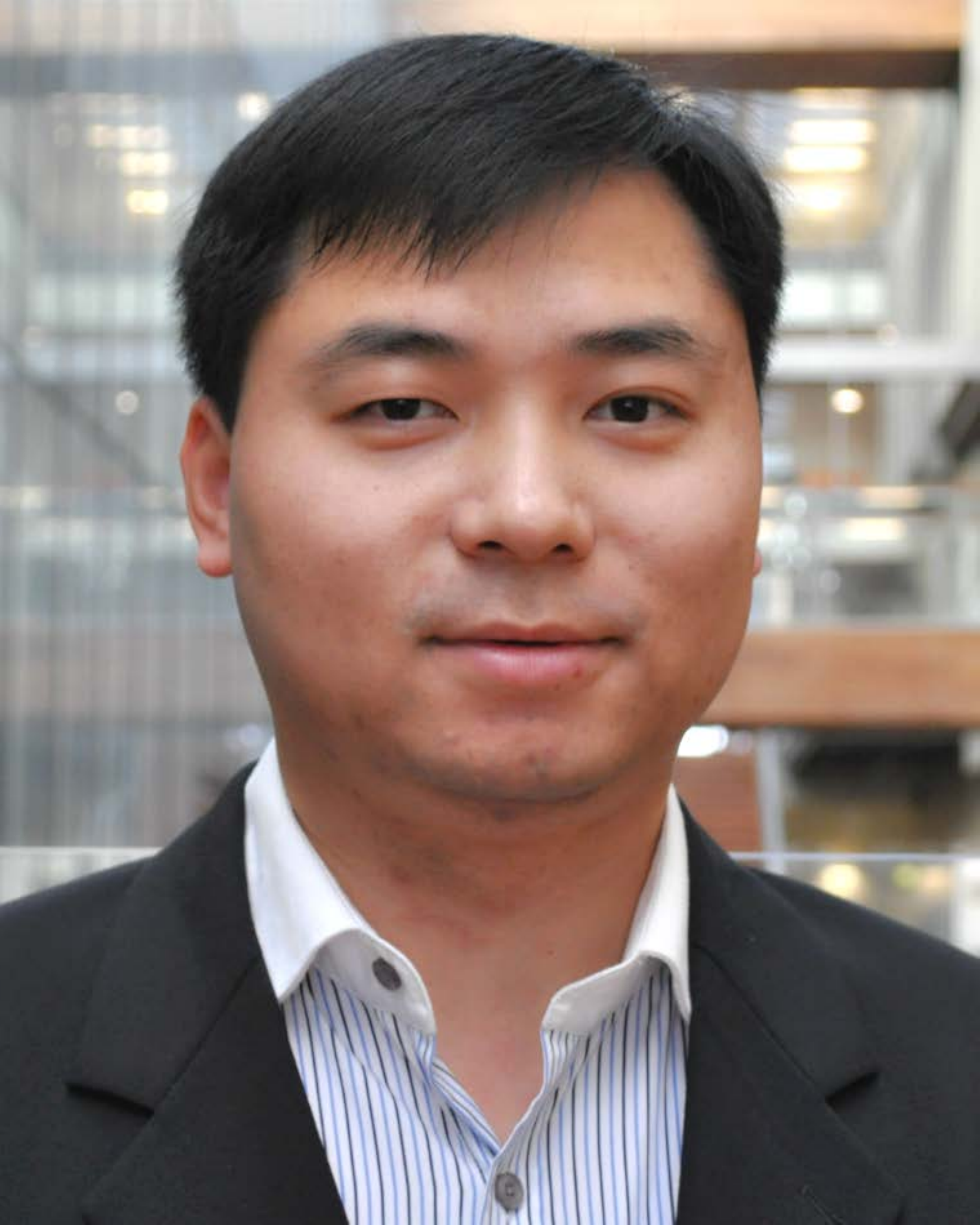}}]{Dacheng Tao} (F'15) is Professor of Computer Science and ARC Laureate Fellow in the School of Computer Science and the Faculty of Engineering and Information Technologies, and the Inaugural Director of the UBTECH Sydney Artificial Intelligence Centre, at the University of Sydney. He mainly applies statistics and mathematics to Artificial Intelligence and Data Science. His research results have expounded in one monograph and 200+ publications at prestigious journals and prominent conferences, such as IEEE T-PAMI, T-IP, T-NNLS,T-CYB, IJCV, JMLR, NIPS, ICML, CVPR, ICCV, ECCV, ICDM; and ACM SIGKDD, with several best paper awards, such as the best theory/algorithm paper runner up award in IEEE ICDM'07, the best student paper award in IEEE ICDM'13, the 2014 ICDM 10-year highest-impact paper award, the 2017 IEEE Signal Processing Society Best Paper Award, and the distinguished paper award in the 2018 IJCAI. He received the 2015 Austrlian Scopus-Eureka Prize and the 2018 IEEE ICDM Research Contributions Award. He is a Fellow of the Australian Academy of Science, AAAS, IEEE, IAPR, OSA and SPIE.
\end{IEEEbiography}

\end{document}